\RequirePackage{fix-cm}
\documentclass[twocolumn]{svjour3}          
\smartqed  

\usepackage[utf8]{inputenc} 
\usepackage[T1]{fontenc}    
\usepackage{url}            
\usepackage{booktabs}       
\usepackage{amsfonts}       
\usepackage{nicefrac}       
\usepackage{microtype}      
\usepackage{xcolor}         
\usepackage{epsfig}
\usepackage{graphicx}
\usepackage{amsmath}
\usepackage{amssymb}
\usepackage{multirow}
\usepackage{color, colortbl}
\usepackage{subcaption}
\usepackage{pifont}
\newcommand{\cmark}{\ding{51}}

\definecolor{COLOR_CSID}{HTML}{e0f5ff}
\definecolor{COLOR_NEAROOD}{HTML}{ffefe0}
\definecolor{COLOR_FAROOD}{HTML}{ffdebf}
\definecolor{COLOR_MEAN}{HTML}{f0f0f0}

\usepackage{xspace}
\usepackage{comment}
\usepackage{bm}
\usepackage{wrapfig}
\usepackage[pagebackref=false, breaklinks=true,letterpaper=true, colorlinks,citecolor=citecolor, linkcolor=linkcolor, bookmarks=false]{hyperref}
\definecolor{citecolor}{HTML}{0071BC}
\definecolor{linkcolor}{HTML}{ED1C24}
\newcommand{\subfig}[1]{\textcolor{linkcolor}{-#1}}

\makeatletter
\renewcommand\paragraph{
  \@startsection{paragraph} 
  {4} 
  {\z@} 
  {.5em \@plus1ex \@minus.2ex} 
  {-1.5em} 
  {\normalfont\normalsize\bfseries} 
}
\DeclareRobustCommand\onedot{\futurelet\@let@token\@onedot}
\def\@onedot{\ifx\@let@token.\else.\null\fi\xspace}

\def\eg{\emph{e.g}\onedot} 
\def\ie{\emph{i.e}\onedot}

\def\etal{\emph{et al}\onedot}
\makeatother

\newcommand{\tabstyle}[1]{
  \setlength{\tabcolsep}{#1}
  \centering
}

\begin{document}
\sloppy

\title{Full-Spectrum Out-of-Distribution Detection
}


\author{Jingkang Yang \and
        Kaiyang Zhou  \and
        Ziwei Liu}


\institute{Jingkang Yang \at
              S-Lab, Nanyang Technological University, Singapore \\
              \email{jingkang001@ntu.edu.sg}
           \and
           Kaiyang Zhou \at
              S-Lab, Nanyang Technological University, Singapore \\
              \email{kaiyang.zhou@ntu.edu.sg}           
           \and
           Ziwei Liu \at
           S-Lab, Nanyang Technological University, Singapore \\
           \email{ziwei.liu@ntu.edu.sg}
}
\date{Received: date / Accepted: date}

\maketitle
\begin{abstract}
Existing out-of-distribution (OOD) detection literature clearly defines semantic shift as a sign of OOD but does not have a consensus over covariate shift. Samples experiencing covariate shift but not semantic shift are either excluded from the test set or treated as OOD, which contradicts the primary goal in machine learning---being able to generalize beyond the training distribution. In this paper, we take into account both shift types and introduce full-spectrum OOD (FS-OOD) detection, a more realistic problem setting that considers both detecting semantic shift and being tolerant to covariate shift; and designs three benchmarks. These new benchmarks have a more fine-grained categorization of distributions (\ie, training ID, covariate-shifted ID, near-OOD, and far-OOD) for the purpose of more comprehensively evaluating the pros and cons of algorithms. To address the FS-OOD detection problem, we propose SEM, a simple feature-based semantics score function. SEM is mainly composed of two probability measures: one is based on high-level features containing both semantic and non-semantic information, while the other is based on low-level feature statistics only capturing non-semantic image styles. With a simple combination, the non-semantic part is cancelled out, which leaves only semantic information in SEM that can better handle FS-OOD detection. Extensive experiments on the three new benchmarks show that SEM significantly outperforms current state-of-the-art methods.
Our code and benchmarks are released in \url{https://github.com/Jingkang50/OpenOOD}.
\end{abstract}
\section{Introduction}
\label{S:intro}

State-of-the-art deep neural networks are notorious for their overconfident predictions on out-of-distribution (OOD) data~\cite{yang2021generalized}, defined as those not belonging to in-distribution (ID) classes. Such a behavior makes real-world deployments of neural network models untrustworthy and could endanger users involved in the systems. To solve the problem, various OOD detection methods have been proposed in the past few years~\cite{baseline,odin,energyood,mahalanobis,likelihood,duq20icml,gram20icml}. The main idea for an OOD detection algorithm is to assign to each test image a score that can represent the likelihood of whether the image comes from in- or out-of-distribution. Images whose scores fail to pass a threshold are rejected, and the decision-making process should be transferred to humans for better handling.

A critical problem in existing research of OOD detection is that only semantic shift is considered in the detection benchmarks while covariate shift---a type of distribution shift that is mainly concerned with changes in appearances like image contrast, lighting or viewpoint---is either excluded from the evaluation stage or simply treated as a sign of OOD~\cite{yang2021generalized}, which contradicts with the primary goal in machine learning, \ie, to generalize beyond the training distribution~\cite{zhou2021domain}.

\begin{figure}
\centering
\includegraphics[width=\linewidth]{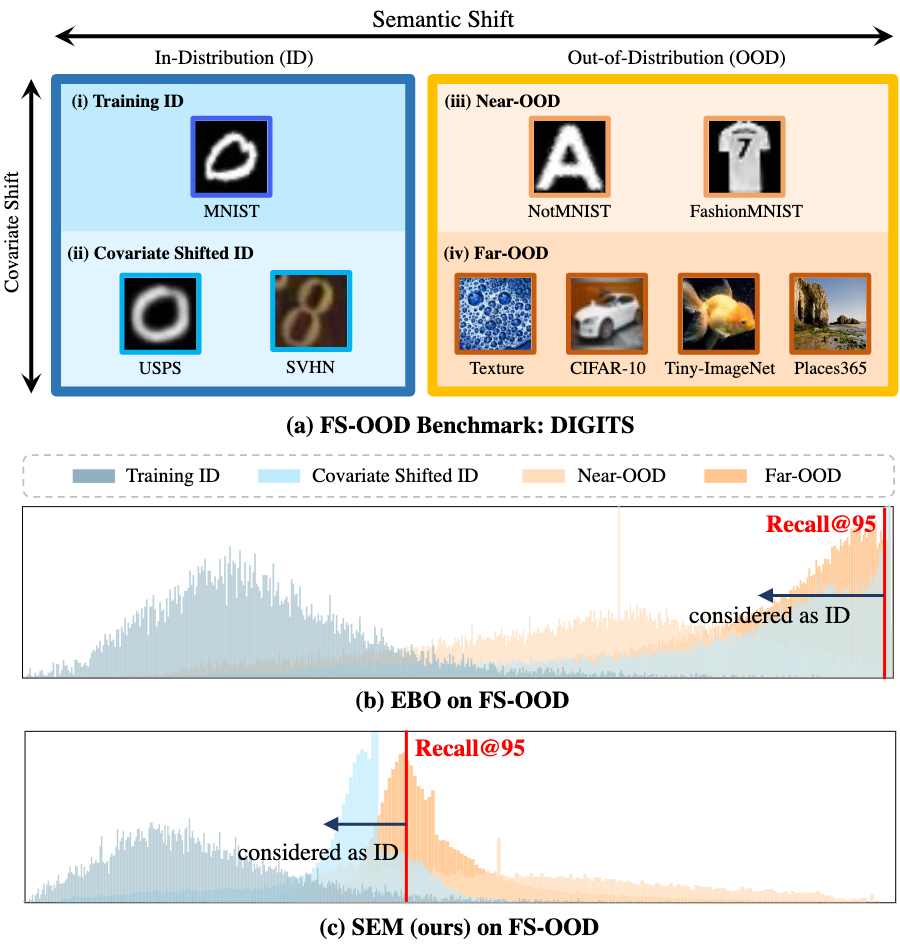}
\vspace{-0.5cm}
\caption{
\textbf{Comparison of OOD detection scores obtained by different approaches on the newly introduced full-spectrum OOD detection benchmark: (a) DIGITS Benchmark}. Ideally, the scores should be clear enough to separate out OOD data while include covariate-shifted data as in-distribution---which has been ignored by most existing research. (b) The state-of-the-art energy-based approach, EBO~\cite{energyood}, apparently fails in this scenario. (c) Our approach, based on a semantics-oriented score function, can improve the detection performance significantly.
}
\label{fig:intro}
\end{figure}

In this paper, we introduce a more challenging yet realistic problem setting called \emph{full-spectrum out-of-distribution detection}, or \emph{FS-OOD detection}. The new setting takes into account both the detection of semantic shift and the ability to recognize covariate-shifted data as ID. To this end, we design three benchmarks, namely DIGITS, OBJECTS and COVID, each targeting a specific visual recognition task and together constituting a comprehensive testbed. We also provide a more fine-grained categorization of distributions for the purpose of thoroughly evaluating an algorithm. Specifically, we divide distributions into four groups: training ID, covariate-shifted ID, near-OOD, and far-OOD (the latter two are inspired by a recent study~\cite{ming2021impact}). Figure~\ref{fig:intro}\subfig{a} shows example images from the DIGITS benchmark: the covariate-shifted images contain the same semantics as the training images, \ie, digits from 0 to 9, and should be classified as ID, whereas the two OOD groups clearly differ in semantics but represent two different levels of covariate shift.

Ideally, an OOD detection system is expected to produce high scores for samples from the training ID and covariate-shifted ID groups, while assign low scores to samples from the two OOD groups. However, when applying a state-of-the-art OOD detection method, \eg the energy-based EBO~\cite{energyood}, to the proposed benchmarks like DIGITS (see Figure~\ref{fig:intro}\subfig{b}), we observe that the resulting scores completely fail to distinguish between ID and OOD. As shown in Figure~\ref{fig:intro}\subfig{b}, all data are classified as ID including both near-OOD and far-OOD samples.

To address the more challenging but realistic FS-OOD detection problem, we propose SEM, \emph{a simple feature-based semantics score function}. Unlike existing score functions that are based on either marginal distribution~\cite{energyood} or predictive confidence~\cite{baseline}, SEM leverages features from both top and shallow layers to deduce a single score that is only relevant to semantics, hence more suitable for identifying semantic shift while ensuring robustness under covariate shift. Specifically, SEM is mainly composed of two probability measures: one is based on high-level features containing both semantic and non-semantic information, while the other is based on low-level feature statistics only capturing non-semantic image styles. With a simple combination, the non-semantic part is cancelled out, which leaves only semantic information in SEM. Figure~\ref{fig:intro}\subfig{c} illustrates that SEM's scores are much clearer to distinguish between ID and OOD.

We summarize the \textbf{contributions} of this paper as follows. \textbf{1)} For the first time, we introduce the full-spectrum OOD detection problem, which represents a more realistic scenario considering both semantic and covariate shift in the evaluation pipeline. \textbf{2)} Three benchmark datasets are designed for research of FS-OOD detection. They cover a diverse set of recognition tasks and have a detailed categorization over distributions. \textbf{3)} A simple yet effective OOD detection score function called SEM is proposed. Through extensive experiments on the three new benchmarks, we demonstrate that SEM significantly outperforms current state-of-the-art methods in FS-OOD detection. The source code and new datasets are open-sourced in 
\url{https://github.com/Jingkang50/OpenOOD}.

\section{Related Work}

The key idea in out-of-distribution (OOD) detection is to design a metric, known as score function, to assess whether a test sample comes from in- or out-of-distribution. The most commonly used metric is based on the conditional probability $p(y|\bm{x})$. An early OOD detection method is maximum softmax probability (MSP)~\cite{baseline}, which is motivated by the observation that deep neural networks tend to give lower confidence to mis-classified or OOD data. A follow-up work ODIN~\cite{odin} applies a temperature scaling parameter to soften the probability distribution, and further improves the performance by injecting adversarial perturbations to the input. Model ensembling has also been found effective in enhancing robustness in OOD detection~\cite{waic,eloc}.

Another direction is to design the metric in a way that it reflects the marginal probability $p(\bm{x})$. Liu \etal~\cite{energyood} connect their OOD score to the marginal distribution using an energy-based formulation, which essentially sums up the prediction logits over all classes. Lee \etal~\cite{mahalanobis} assume the source data follow a normal distribution and learn a Mahalanobis distance to compute the discrepancy between test images and the estimated distribution parameters. Generative modeling has also been investigated to estimate a likelihood ratio for scoring test images~\cite{waic,likelihood,S}.

Some methods exploit external OOD datasets. For example, Hendrycks \etal~\cite{hendrycks18oe} extend MSP by training the model to produce uniform distributions on external OOD data. Later works introduce re-sampling strategy~\cite{backgroundsample} and cluster-based methodology~\cite{yang2021scood} to better leverage the background data. However, this work do not use external OOD datasets for model design.

Different from all existing methods, our approach aims to address a more challenging scenario, \ie, FS-OOD detection, which has not been investigated in the literature but is critical to real-world applications. The experiments show that current state-of-the-art methods mostly fail in the new setting while our approach gains significant improvements.
\section{Methodology}
\label{S:method}

\begin{figure*}[t]
    \centering
    \includegraphics[width=0.9\linewidth]{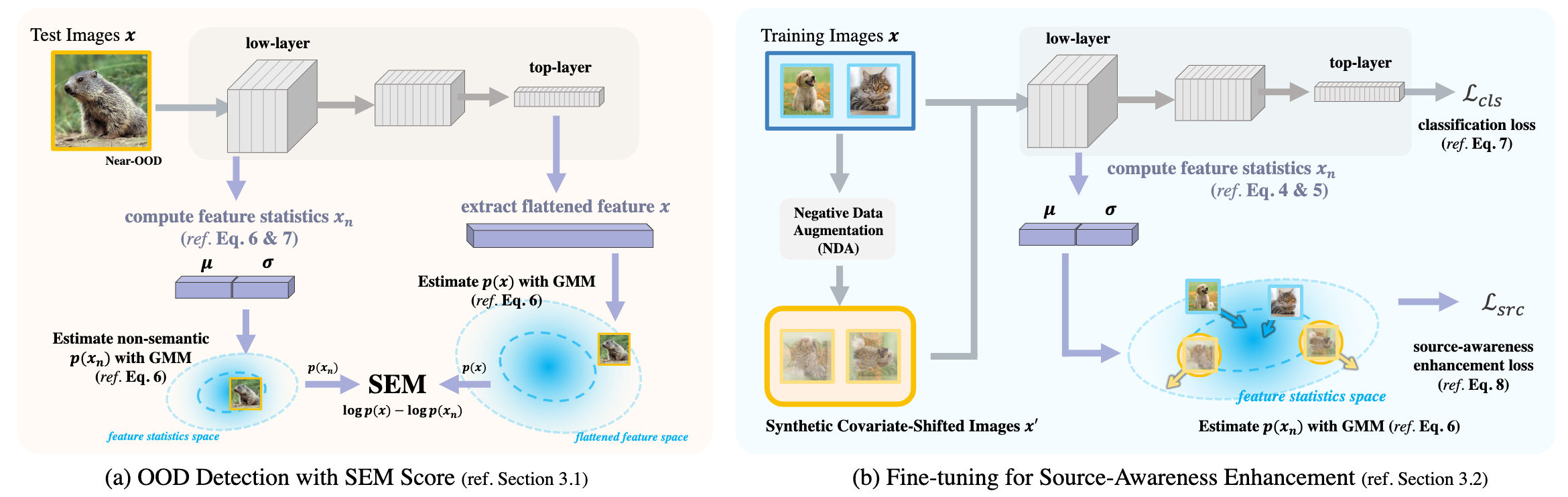}
    \vspace{-3pt}
    \caption{\textbf{Overview of our Methodology.}
    (a) The computation of SEM score function for OOD detection. SEM combines the estimation of $p(\bm{x})$ (using top-layer features to capture both semantic and non-semantic information) and $p(\bm{x}_n)$ (using low-level feature statistics to only capture non-semantic information) with Eq.~\ref{eq:final_score} for better concentration on semantics.
    (b) The fine-tuning scheme to enhance source-awareness for better estimating $p(\bm{x}_n)$. The main idea is to pull together the instance-level CNN feature statistics of in-distribution data to make them more compact, while pushing away those of synthetic OOD data, which are obtained by negative data augmentation such as Mixup~\cite{mixup}.}
    \label{fig:enhance_source_awareness}
    \vspace{-10pt}
\end{figure*}

\subsection{Feature-Based Semantics Score Function}
\label{sec:method;subsec:score}

Key to detect out-of-distribution (OOD) data lies in the design of a score function, which is used as a quantitative measure to distinguish between in- and out-of-distribution data. Our idea is to design the function in such a way that the degree of semantic shift is effectively captured, \ie, the designed score to be only sensitive to semantic shift while being robust to covariate shift. For data belonging to the in-distribution classes, the score is high, and vice versa.

\paragraph{Formulation}
Our score function, called SEM, has the following design:
\begin{equation} \label{eq:score_init}
\operatorname{SEM}(\bm{x}) = \log p(\bm{x}_s),
\end{equation}
where $\bm{x}$ denotes image features learned by a neural network; and $\bm{x}_s$ denotes features that only capture the semantics. The probability $p(\bm{x}_s)$ can be computed by a probabilistic model, such as a Gaussian mixture model.

The straightforward way to model $\bm{x}_s$ is to learn a neural network for image recognition and hope that the output features $\bm{x}$ only contain semantic information, \ie, $\bm{x}_s = \bm{x}$. If so, the score can be simply computed by $\operatorname{SEM}(\bm{x}) = \log p(\bm{x})$. However, numerous studies have suggested that the output features $\bm{x}$ often contain both semantic and non-semantic information while decoupling them is still an open research problem~\cite{zhou2021domain,lin2021domain,peng2019domain}. Let $\bm{x}_n$ denote non-semantic features, we assume that semantic features $\bm{x}_s$ and non-semantic features $\bm{x}_n$ are generated independently, namely
\begin{equation} \label{eq:decom}
p(\bm{x}) = p(\bm{x}_s) p(\bm{x}_n).
\end{equation}

We propose a simple method to model the score function so that it becomes only relevant to the semantics of an image. This is achieved by leveraging \emph{low-level feature statistics}, \ie, means and standard deviations, learned in a CNN, which have been shown effective in capturing image styles that are essentially irrelevant to semantics~\cite{mixstyle}. Specifically, the score function in Eq.~\ref{eq:score_init} is rewritten as
\begin{align}
\operatorname{SEM}(\bm{x}) = \log p(\bm{x}_s) = \log \frac{p(\bm{x}_s) p(\bm{x}_n)}{p(\bm{x}_n)} = \log \frac{p(\bm{x})}{p(\bm{x}_n)}, \label{eq:final_score}
\end{align}
where $p(\bm{x})$ is computed using the output features while $p(\bm{x}_n)$ is based on low-level feature statistics.

Below we first discuss how to compute feature statistics and then detail the approach of how to model the distributions for $\bm{x}$ and $\bm{x}_n$.

\paragraph{Feature Statistics Computation}
Instance-level feature statistics have been widely used in the style transfer community for manipulating image style~\cite{adain}. Given a set of CNN feature maps $\bm{z} \in \mathbb{R}^{C \times H \times W}$ with $C$, $H$ and $W$ denoting the number of channels, height and width, their feature statistics, \ie, means $\bm{\mu} \in \mathbb{R}^{C}$ and standard deviations $\bm{\sigma} \in \mathbb{R}^{C}$, are computed across the spatial dimension within each channel $c = \{1, 2, ..., C\}$,
\begin{align}
\mu_c &= \frac{1}{HW} \sum_{h=1}^H \sum_{w=1}^W z_{c, h, w}, \\
\sigma_c &= \left(\frac{1}{HW} \sum_{h=1}^H \sum_{w=1}^W (z_{c, h, w} - \mu_c)^2\right)^{\frac{1}{2}}.
\end{align}

As shown in Zhou \etal~\cite{mixstyle}, the feature statistics in shallow CNN layers are strongly correlated with domain information (\ie, image style) while those in higher layers pick up more semantics. Therefore, we choose to extract feature statistics in the first CNN layer and represent $\bm{x}_n$ by concatenating the means and standard deviations, \ie, $\bm{x}_n = [\bm{\mu}, \bm{\sigma}]^T$.

\paragraph{Distribution Modeling}
For simplicity, we model $p(\bm{x})$ and $p(\bm{x}_n)$ in Eq.~\ref{eq:final_score} using the same approach, which consists of two steps: dimension reduction and distribution modeling. Below we only discuss $p(\bm{x})$ for clarity.

Motivated by the manifold assumption in Bengio \etal~\cite{bengio2013representation} that suggests data typically lie in a manifold of much lower dimension than the input space, we transform features $\bm{x}$ to a new low-dimensional space, with a hope that the structure makes it easier to distinguish between in- and out-of-distribution. To this end, we propose a variant of the principal component analysis (PCA) approach. Specifically, rather than maximizing the variance for the entire population, we maximize the sum of variances computed within each class with respect to the transformation matrix. In doing so, we can identify a space that is less correlated with classes.

Given a training dataset, we build a Gaussian mixture model (GMM) to capture $p(\bm{x})$. Formally, $p(\bm{x})$ is defined as
\begin{equation} \label{eq:gmm}
p(\bm{x}) = \sum_{m=1}^M \lambda_m \mathcal{N}(\bm{\alpha}_m, \bm{\beta}_m),
\end{equation}
where $M$ denotes the number of mixture components, $\lambda_m$ the mixture weight s.t.~$\sum_{m=1}^M \lambda_m = 1$, and $\bm{\alpha}_m$ and $\bm{\beta}_m$ the means and variances of a normal distribution. A GMM model can be efficiently trained by the expectation-maximization (EM) algorithm.

\subsection{Source-Awareness Enhancement}
\label{sec:method;subsec:enhance}

While feature statistics exhibit a higher correlation with source distributions~\cite{mixstyle}, the boundary between in- and out-of-distribution in complicated real-world data is not guaranteed to be clear enough for differentiation. Inspired by Liu et al.~\cite{energyood} who fine-tune a pretrained model to increase the energy values assigned to OOD data and lower down those for ID data, we propose a fine-tuning scheme to enhance source-awareness in feature statistics. An overview of the fine-tuning scheme is illustrated in Figure~\ref{fig:enhance_source_awareness}\subfig{b}.

\paragraph{Negative Data Augmentation}
The motivation behind our fine-tuning scheme is to obtain a better estimate of non-semantic score, in hope that it will help SEM better capture the semantics with the combination in Eq.~\ref{eq:final_score}.
This can be achieved by explicitly training feature statistics of ID data to become more compact, while pushing OOD data's feature statistics away from the ID support areas. A straightforward way is to collect auxiliary OOD data like Liu \etal~\cite{energyood} for building a contrastive objective. In this work, we propose a more efficient way by using negative data augmentation~\cite{sinha2021negative} to synthesize OOD samples. The key idea is to choose data augmentation methods to easily generate samples with covariate shift. One example augmentation is Mixup~\cite{mixup}.

\paragraph{Learning Objectives}
Given a source dataset $\mathcal{S} = \{(\bm{x}, y)\}$,\footnote{With a slight abuse of notation, we use $\bm{x}$ here to denote an image.} we employ negative data augmentation methods $\operatorname{aug}(\cdot)$ to synthesize an OOD dataset $\mathcal{S}_{aug} = \{(\bm{x}', y)\}$ where $\bm{x}' = \operatorname{aug}(\bm{x})$. For fine-tuning, we combine a classification loss $\mathcal{L}_{cls}$ with a source-awareness enhancement loss $\mathcal{L}_{src}$. These two losses are formally defined as
\begin{equation} \label{eq:losses}
\mathcal{L}_{cls} = - \sum_{(\bm{x}, y) \sim \mathcal{S}} \log p(y|\bm{x}), 
\end{equation}
and
\begin{equation}
\mathcal{L}_{src} = \sum_{\bm{x}' \sim \mathcal{S}_\text{aug}} p(\bm{x}_n') - \sum_{\bm{x} \sim \mathcal{S}} p(\bm{x}_n),
\end{equation}
where the marginal probability $p(\bm{x})$ is computed based on a GMM model described previously. Note that the GMM model is updated every epoch to adapt to the changing features.

After fine-tuning, we learn a new GMM model using the original source dataset. This model is then used to estimate the marginal probability $p(\bm{x})$ at test time.

\section{FS-OOD Benchmarks}
\label{S:benchmark}

\begin{figure}
    \centering
    \includegraphics[width=\linewidth]{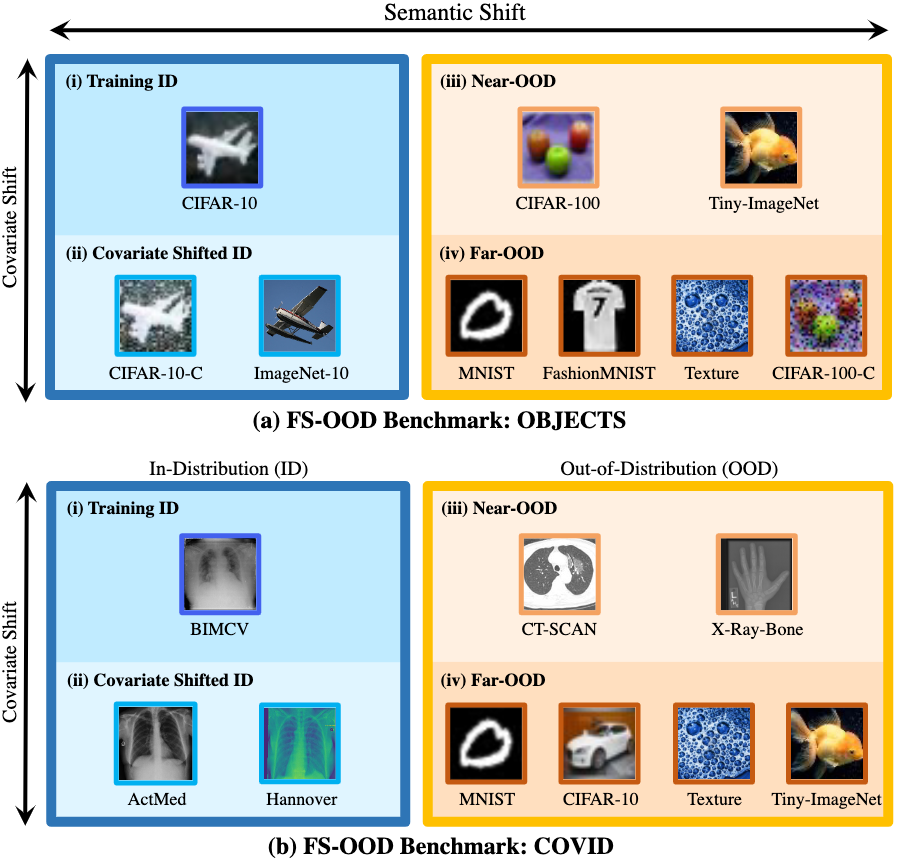}
    \caption{\textbf{Examples for the two FS-OOD detection benchmarks: COVID and OBJECTS}.
    Each benchmark consists of a training ID dataset, two covariate-shifted ID datasets, two near-OOD datasets, and four far-OOD datasets.
    }
    \vspace{-10pt}
    \label{fig:benchmark}
\end{figure}

To evaluate full-spectrum out-of-distribution (FS-OOD) detection algorithms, we design three benchmarks: DIGITS, OBJECTS, and COVID. Examples for DIGITS are shown in Figure~\ref{fig:intro} and the other two are shown in Figure~\ref{fig:benchmark}.


\paragraph{Benchmark-1: DIGITS}
We construct the DIGITS benchmark based on the popular digit datasets: MNIST~\cite{mnist}, which contains 60,000 images for training. During testing, the model will be exposed to 10,000 MNIST test images, with 26,032 covariate-shifted ID images from SVHN~\cite{svhn} and another 9,298 from USPS~\cite{usps}. The near-OOD datasets are notMNIST~\cite{notmnist} and FashionMNIST~\cite{fashionmnist}, which share a similar background style with MNIST. The far-OOD datasets consist of a textural dataset (Texture~\cite{texture}), two object datasets (CIFAR-10~\cite{cifar} \& Tiny-ImageNet~\cite{imagenet}), and one scene dataset (Places365~\cite{places365}). The CIFAR-10 and Tiny-ImageNet test sets have 10,000 images for each. The Places365 test set contains 36,500 scene images.

\paragraph{Benchmark-2: OBJECTS}
The OBJECTS benchmark is built on top of CIFAR-10~\cite{cifar}, which contains 50,000 images for training. During testing, the model will be exposed to 10,000 CIFAR-10 test images, and another 10,000 images selected from ImageNet-22K~\cite{imagenet} with the same categories as CIFAR-10 (so it is called ImageNet-10). For ImageNet-10, we choose five ImageNet-22K classes corresponding to one CIFAR-10 class, with each class selecting 1,000 training images and 200 testing images. Details of the selected classes are shown in Table~\ref{T:imagenet22k}.
In addition to ImageNet, CIFAR-10-C is used as a covariate-shifted ID dataset, which is essentially a corrupted version of CIFAR-10.
For near-OOD, we choose CIFAR-100 and Tiny-ImageNet. For far-OOD, we choose MNIST, FashionMNIST, Texture and CIFAR-100-C.

\paragraph{Benchmark-3: COVID}
We construct a real-world benchmark to show the practical value of FS-OOD. We simulate the scenario where an AI-assisted diagnostic system is trained to identify COVID-19 infection from chest x-ray images. The training data come from a single source (\eg, a hospital) while the covariate-shifted ID test data are from other hospitals or machines, to which the system needs to be robust and produce reliable predictions. Specifically, we refer to the COVID-19 chest X-ray dataset review~\cite{santa2021public}, and use the large-scale image collection from Valencian Region Medical ImageBank~\cite{vaya2020bimcv} (referred to as BIMCV) as training ID images (randomly sampled 2443 positive cases and 2501 negative cases with necessary cleaning). Images from two other sources, \ie, ACTUALMED~\cite{Wang2020} (referred to as ActMed with 132 positive images), and Hannover~\cite{winther12275009covid} (from Hannover Medical School with 243 positive images), are considered as the covariate-shifted ID group. OOD images are from completely different classes. Near-OOD images are obtained from other medical datasets, \ie, the RSNA Bone Age dataset with 200 bone X-ray images~\cite{boneage} and 544 COVID CT images~\cite{covidct}. Far-OOD samples are defined as those with drastic visual and concept differences than the ID images. We use MNIST, CIFAR-10, Texture and Tiny-ImageNet.

\begin{table}[]
\caption{\textbf{Selected ImageNet-22K classes for OBJECTS benchmark.} 
We manually find 5 ImageNet-22K classes that belong to each CIFAR-10 classes,
and pick the first 1,000 images from every selected class for OBJECTS benchmark.
A string such as `n03365231' is the synset id for downloading the corresponding class from ImageNet API.}
\label{T:imagenet22k}
\centering
\resizebox{\linewidth}{!}{
\begin{tabular}{l|l|l|l}
\toprule
\textbf{Airplane} & \textbf{Automobile} & \textbf{Bird} & \textbf{Cat} \\ 
\midrule
\begin{tabular}[c]{@{}l@{}}
n03365231 floatplane\\
n02691156 airplane\\
n04552348 warplane\\
n02686568 aircraft\\
n02690373 airliner\\
\end{tabular}
&
\begin{tabular}[c]{@{}l@{}}
n04516354 used car\\
n04285008 sports car\\
n02958343 car\\
n03594945 jeep\\
n02930766 cab\\
\end{tabular}
&
\begin{tabular}[c]{@{}l@{}}
n01503061 bird\\
n01812337 dove\\
n01562265 robin\\
n01539573 sparrow\\
n01558594 blackbird\\
\end{tabular}
&
\begin{tabular}[c]{@{}l@{}}
n02121808 domestic cat\\
n02123159 tiger cat\\
n02122878 tabby\\
n02123394 Persian cat\\
n02123597 Siamese cat\\
\end{tabular}\\
\bottomrule
\end{tabular}
}
\vspace{0.3cm}

\resizebox{\linewidth}{!}{
\begin{tabular}{l@{\hskip 24pt}|l@{\hskip 24pt}|l@{\hskip 24pt}}
\toprule
\textbf{Deer} & \textbf{Dog} & \textbf{Frog} \\ 
\midrule
\begin{tabular}[c]{@{}l@{}}
n02430045 deer\\
n02431122 red deer\\
n02432511 mule deer\\
n02433318 fallow deer\\
n02431976 Japanese deer\\
\end{tabular}
&
\begin{tabular}[c]{@{}l@{}}
n02116738 African hunting dog\\
n02087122 hunting dog\\
n02105855 Shetland sheepdog\\
n02109961 Eskimo dog\\
n02099601 golden retriever\\
\end{tabular}
&
\begin{tabular}[c]{@{}l@{}}
n01639765 frog\\
n01641577 bullfrog\\
n01644373 tree frog\\
n01640846 true frog\\
n01642539 grass frog\\
\end{tabular}\\
\bottomrule
\end{tabular}
}
\vspace{0.3cm}

\resizebox{\linewidth}{!}{
\begin{tabular}{l@{\hskip 24pt}|l@{\hskip 24pt}|l@{\hskip 24pt}}
\toprule
\textbf{Horse} & \textbf{Ship} & \textbf{Truck} \\ 
\midrule
\begin{tabular}[c]{@{}l@{}}
n02387254 farm horse\\
n02381460 wild horse\\
n02374451 horse\\
n02382948 racehorse\\
n02379183 quarter horse\\
\end{tabular}
&
\begin{tabular}[c]{@{}l@{}}
n02965300 cargo ship\\
n04194289 ship\\
n03095699 container ship\\
n02981792 catamaran\\
n03344393 fireboat\\
\end{tabular}
&
\begin{tabular}[c]{@{}l@{}}
n04490091 truck\\
n03417042 garbage truck\\
n03173929 delivery truck\\
n04467665 trailer truck\\
n03345487 fire engine\\
\end{tabular}\\
\bottomrule
\end{tabular}
}
\end{table}

\paragraph{Evaluation Metrics}
In the FS-OOD setting, different datasets belonging to one OOD type (\ie, near-OOD or far-OOD) are grouped together. We also report the performance on contrasting covariate-shifted ID with training ID, although covariate-shifted ID are not OOD samples.
We use three metrics to evaluate the OOD detection performance, which are detailed as follows: \textbf{1) FPR95} stands for false positive rate measured when true positive rate (TPR) sits at 95\%. Intuitively, FPR95 measures the portion of samples that are falsely recognized as in-distribution data when most true in-distribution samples are recalled. \textbf{2) AUROC} refers to the Area Under the Receiver Operating Characteristic curve, which is concerned with both FPR and TPR. \textbf{3) AUPR} means the Area Under the Precision-Recall curve, which considers both precision and recall. For FPR95, the lower the value, the better the model. For AUROC and AUPR, the higher the value, the better the model.

\begin{table*}
    \tabstyle{5pt}
	\caption{\textbf{Comparison between previous state-of-the-art methods and the proposed SEM score on FS-OOD benchmarks.}
	\label{T:main_result}
	The proposed SEM obtains a consistently better performance on most of the metrics than
	MSP~\cite{baseline}, ODIN~\cite{odin}, Energy-based OOD (EBO) score~\cite{energyood}, and Mahalanobis Distance Score (MDS)~\cite{mahalanobis}, especially on the near-OOD scenarios.}
	\vspace{-5pt}
	\centering
	\resizebox{0.95\textwidth}{!}{
	\begin{tabular}{@{\hskip 8pt}l@{\hskip 6pt}|@{\hskip 6pt}ccccc@{\hskip 6pt}|@{\hskip 6pt}ccccc@{\hskip 6pt}|@{\hskip 6pt}ccccc@{\hskip 8pt}}
		\toprule
		& \multicolumn{5}{c@{\hskip 6pt}|@{\hskip 6pt}}{FPR95~$\downarrow$} 
		& \multicolumn{5}{c@{\hskip 6pt}|@{\hskip 6pt}}{AUROC~$\uparrow$} 
		& \multicolumn{5}{c@{\hskip 6pt}}{AUPR~$\uparrow$} \\ 
		& MSP & ODIN & EBO & MDS & SEM & MSP & ODIN & EBO & MDS & SEM & MSP & ODIN & EBO & MDS & SEM \\
		\midrule
		\multicolumn{16}{l}{\textbf{- DIGITS} 
		(Training ID: \underline{MNIST}, Covariate-Shifted ID: \underline{USPS \& SVHN})} \vspace{.1cm} \\
		\rowcolor{COLOR_NEAROOD}
		notmnist
        & 99.97& 99.95& 99.99& 78.83& \textbf{10.93} 
        & 32.54& 29.04& 25.49& 79.10& \textbf{96.74} 
        & 67.33& 65.97& 63.97& 90.60& \textbf{98.54} \\
		\rowcolor{COLOR_NEAROOD}
	    FashionMNIST
	    & 99.90& 99.97& 99.98& 94.68& \textbf{68.63} 
        & 39.71& 38.51& 37.64& 60.42& \textbf{80.20} 
        & 82.40& 82.16& 81.57& 88.84& \textbf{94.38} \\
        \midrule
        \rowcolor{COLOR_MEAN}
	    Mean (Near-OOD) 
	    & 99.93& 99.96& 99.98& 86.75& \textbf{39.78} 
        & 36.12& 33.77& 31.56& 69.76& \textbf{88.47} 
        & 74.87& 74.06& 72.77& 89.72& \textbf{96.46} \\
		\midrule
		\rowcolor{COLOR_FAROOD}
		Texture  
		& 94.89& 94.65& 98.40& \textbf{87.46}& 90.90 
        & 64.34& 64.02& 65.02& 72.42& \textbf{74.45} 
        & 94.40& 94.32& 94.47& 95.81& \textbf{96.12} \\
		\rowcolor{COLOR_FAROOD}
		CIFAR-10
        & 98.01& 98.38& 99.62& 95.47& \textbf{91.57} 
        & 52.22& 51.15& 50.95& 67.96& \textbf{69.29} 
        & 87.26& 86.84& 86.36& 91.74& \textbf{92.06} \\
	    \rowcolor{COLOR_FAROOD}
	    Tiny-ImageNet
        & 97.98& 98.23& 99.58& 96.20& \textbf{93.39} 
        & 52.94& 51.98& 51.89& 64.31& \textbf{67.54} 
        & 87.51& 87.15& 86.72& 90.71& \textbf{91.58} \\
	    \rowcolor{COLOR_FAROOD}
	    Places365
        & 98.68& 98.78& 99.65& 98.06& \textbf{94.15} 
        & 50.22& 49.30& 48.95& 65.42& \textbf{67.63} 
        & 67.11& 66.51& 65.41& 76.64& \textbf{77.61} \\
		\midrule
		\rowcolor{COLOR_MEAN}
		Mean (Far-OOD)  
        & 97.39& 97.51& 99.31& 94.30& \textbf{92.50} 
        & 54.93& 54.11& 54.20& 67.53& \textbf{69.73} 
        & 84.07& 83.71& 83.24& 88.73& \textbf{89.34} \\
		\midrule
		\midrule
		\multicolumn{16}{l}{\textbf{- OBJECTS}
		(Training ID: \underline{CIFAR-10}, Covariate-Shifted ID: \underline{CIFAR-10-C \& ImageNet-10})} \vspace{.1cm} \\
		\rowcolor{COLOR_NEAROOD}
		CIFAR-100
		&89.44	&87.51	&\textbf{83.84}	&86.28	&86.96	
		&70.17	&60.29	&63.85	&72.05	&\textbf{74.70}	
		&88.28	&81.11	&83.51	&89.42	&\textbf{90.64}\\
		\rowcolor{COLOR_NEAROOD}
	    Tiny-ImageNet
	    &88.22	&88.13	&\textbf{81.58}	&87.45	&86.59	
	    &72.92	&62.07	&67.97	&72.94	&\textbf{76.76}	
	    &90.04	&82.49	&86.30	&89.96	&\textbf{91.86}\\
        \midrule
        \rowcolor{COLOR_MEAN}
	    Mean (Near-OOD)
	    & 88.83	&87.82	&\textbf{82.71}	&86.87	&86.77	
	    &71.55	&61.18	&65.91	&72.50	&\textbf{75.73}	
	    &89.16	&81.80	&84.91	&89.69	&\textbf{91.25}\\
		\midrule
		\rowcolor{COLOR_FAROOD}
		MNIST
		&93.54	&\textbf{82.04}	&92.23	&84.59	&99.70	
		&66.98	&70.31	&54.55	&\textbf{77.04}	&75.69	
		&52.66	&49.58	&34.14	&65.31	&\textbf{76.61}\\
		\rowcolor{COLOR_FAROOD}
		FashionMNIST
		&88.08	&\textbf{68.73}	&72.40	&77.17	&93.72	
		&73.78	&\textbf{80.98}	&76.50	&80.33	&79.40	
		&90.15	&91.53	&89.80	&92.28	&\textbf{93.14}\\
	    \rowcolor{COLOR_FAROOD}
	    Texture
	    &85.64	&\textbf{72.91}	&75.57	&72.98	&82.15	
	    &74.18	&70.14	&68.63	&72.02	&\textbf{79.69}	
	    &93.34	&89.97	&89.51	&88.46	&\textbf{95.48}\\
	    \rowcolor{COLOR_FAROOD}
    	CIFAR-100-C
    	&87.26	&84.26	&\textbf{83.64}	&85.53	&83.92	
    	&74.12	&67.51	&68.37	&68.13	&\textbf{78.89}	
    	&89.74	&83.97	&85.54	&82.97	&\textbf{92.07}\\
		\midrule
		\rowcolor{COLOR_MEAN}
		Mean (Far-OOD)
		&88.63	&76.98	&80.96	&\textbf{80.07}	&89.87	
		&72.27	&72.23	&67.01	&74.38	&\textbf{78.42}	
		&81.47	&78.76	&74.75	&82.25	&\textbf{89.33}\\
		\midrule
		\midrule
    	\multicolumn{16}{l}{\textbf{- COVID} 
    	(Training ID: \underline{BIMCV}, Covariate-Shifted ID: \underline{ActMed \& Hannover})}
    	\vspace{.1cm} \\
		\rowcolor{COLOR_NEAROOD}
		CT-SCAN  
		& 99.80 & 93.06 & 97.35 & 99.39 & \textbf{2.24} 
		& 11.31 & 26.57 & 13.14 & 81.21 & \textbf{99.51} 
		& 52.92 & 57.44 & 53.34 & 94.31 & \textbf{99.80} \\
		\rowcolor{COLOR_NEAROOD}
	    XRayBone  
	    & 97.00 & 55.50 & 42.00 & 100.00 & \textbf{14.50} 
	    & 32.08 & 64.73 & 77.80 & 78.72 & \textbf{94.97} 
	    & 76.95 & 86.11 & 91.68 & 96.67 & \textbf{98.95} \\
        \midrule
        \rowcolor{COLOR_MEAN}
	    Mean (Near-OOD) 
	    & 98.40 & 74.28 & 69.67 & 99.69 & \textbf{8.37} 
	    & 21.70 & 45.65 & 45.47 & 79.96 & \textbf{97.24} 
	    & 64.94 & 71.77 & 72.51 & 95.49 & \textbf{99.37} \\
		\midrule
		\rowcolor{COLOR_FAROOD}
		MNIST
		& 98.30 & 65.14 & 0.35 & 100.00 & \textbf{0.00} 
		& 24.89 & 65.37 & 99.91 & 80.81 & \textbf{100.00} 
		& 1.07 & 2.33 & 95.90 & 81.11 & \textbf{100.00} \\
		\rowcolor{COLOR_FAROOD}
		CIFAR-10
		& 96.32 & 84.61 & 94.67 & 98.02 & \textbf{85.58} 
		& 41.12 & 57.70 & 45.23 & 77.05 & \textbf{52.50} 
		& 8.73 & 12.17 & 9.77 & 61.14 & \textbf{11.27} \\
	    \rowcolor{COLOR_FAROOD}
	    Texture
	    & 98.39 & 94.59 & 87.06 & 56.38 & \textbf{27.57} 
	    & 22.63 & 31.13 & 34.95 & 89.84 & \textbf{90.94} 
	    & 11.43 & 12.59 & 13.14 & 85.50 & \textbf{64.71} \\
	    \rowcolor{COLOR_FAROOD}
	    Tiny-ImageNet
	    & 97.78 & 90.26 & 92.73 & 92.11 & \textbf{44.99} 
	    & 30.26 & 42.76 & 32.69 & 81.99 & \textbf{83.42} 
	    & 7.42 & 8.90 & 7.65 & 77.94 & \textbf{31.19} \\
		\midrule
		\rowcolor{COLOR_MEAN}
		Mean (Far-OOD)
		& 97.70 & 83.65 & 68.70 & 86.63 & \textbf{39.54} 
		& 29.73 & 49.24 & 53.20 & 82.42 & \textbf{81.72} 
		& 7.16 & 9.00 & 31.62 & 76.42 & \textbf{51.79} \\
	\bottomrule
	\end{tabular}}
	\vspace{-5pt}
\end{table*}

\section{Experiments}

\paragraph{Implementation Details}
We conduct experiments on the three proposed FS-OOD benchmarks, \ie, DIGITS, OBJECTS, and COVID. In terms of architectures, we use LeNet-5~\cite{lenet} for DIGITS and ResNet-18~\cite{resnet} for both OBJECTS and COVID.
All models are trained by the SGD optimizer with a weight decay of $5\times 10^{-4}$ and a momentum of 0.9.
For DIGITS and OBJECTS, we set the initial learning rate to 0.1, which is decayed by the cosine annealing rule, and the total epochs to 100. For COVID benchmark, the initial learning rate is set to 0.001 and the model is trained for 200 epochs. When fine-tuning for source-awareness enhancement, the learning rate is set to 0.005 and the total number of epochs is 10. The batch size is set to 128 for all benchmarks.

Notice that the baseline implementations of ODIN~\cite{odin} and MDS~\cite{mahalanobis} require validation set for hyperparameter tuning, we spare a certain portion of near-OOD for validation. More specifically, we use 1,000 notMNIST images for the DIGITS benchmark, 1,000 CIFAR-100 images for the OBJECTS benchmark, and 54 images from CT-SCAN dataset for the COVID benchmark. The proposed method SEM relies on the hyperparameter of $M=3$ for low-layer $p(x_n)$ and number of classes for high-layer $p(x)$ in Gaussian mixture model. For output features with dimensions over 50, PCA is performed to reduce the dimensions to 50.

\subsection{Results on FS-OOD Setting}
\label{S:exp_sota}
We first discuss the results on near- and far-OOD datasets. Table~\ref{T:main_result} summarizes the results where the proposed SEM is compared with current state-of-the-art methods including MSP~\cite{baseline}, ODIN~\cite{odin}, Mahalanobis distance score~(MDS), and Energy-based OOD~\cite{energyood}.

\paragraph{DIGITS Benchmark}
For the DIGITS benchmark, SEM gains significant improvements in all metrics (FPR95, AUROC, and AUPR). A huge gain is observed on notMNIST, which is a challenging dataset due to its closeness in background to the training ID MNIST.
While none of the previous softmax/logits-based methods (\eg, MSP, ODIN, and EBO) are capable to solve the notMNIST problem, the proposed SEM largely reduces the FPR95 metric from 99\% to 10.93\%, and the AUROC is increased from around 30\% to beyond 95\%.
One explanation of the clear advantage is that, the previous output-based OOD detection methods largely depend on the covariate shift to detect OOD samples, while the feature-based MDS (partly rely on top-layer semantic-aware features) and the proposed SEM uses more semantic information, which is critical to distinguish MNIST and notMNIST.
In other words, in the MNIST/notMNIST scenario where ID and OOD have high visual similarity, large dependency on covariate shift while ignorance on the semantic information will lead to the failure of OOD separation.
Similar advantages are also achieved with the other near-OOD dataset.

\paragraph{OBJECTS Benchmark}
Similar to DIGITS benchmark, the proposed SEM surpasses the previous state-of-the-art methods on the near-OOD scenario of the OBJECTS benchmark, especially on the more robust metrics of AUROC and AUPR.
However, the performance gap is not as large as DIGITS.
One explanation is that images in OBJECTS benchmark are more complex than DIGITS, leading the neural networks to be more semantics-orientated.
Therefore, more semantic information is encoded in the previous output-based methods. 
Nevertheless, the proposed SEM method still outperforms others on most of the metrics.
We also notice that SEM score does not reach the best performance on MNIST and FashionMNIST.
One explanation is that two black-and-white images in these two datasets inherently contain significant covariate shifts comparing to both training ID and covariate-shifted ID, so that the scores that efficient on covariate shift detection (\eg, ODIN) can also achieve good results on these datasets. However, these methods fail in near-OOD scenario, as they might believe CIFAR-10-C should be more likely to be OOD than CIFAR-100.

\paragraph{COVID Benchmark}
In this new and real-world application of OOD detection, the proposed SEM score achieves an extraordinary performance on all metrics, which surpasses the previous state-of-the-art methods by a large margin in both near and far-OOD scenarios. 
The result also indicates that previous output-based methods generally breaks down on this setting, \eg, their FPR@95 scores are generally beyond 90\% in near-OOD setting which means ID and OOD are totally mixed. However, the proposed SEM achieves around 10\% in near-OOD setting. On far-OOD samples, the output-based methods are still unable to be sensitive to the ID/OOD discrepancy. The phenomenon matches the performance in DIGITS dataset, where the training data is simple and the logits might learn much non-semantic knowledge to be cancelled out.

\begin{table*}
    \tabstyle{5pt}
	\caption{\textbf{Comparison between previous state-of-the-art methods, the proposed SEM score, and the low-level probabilistic component $p(\bm{x}_n)$ on classic OOD benchmarks, without the existence of covariate-shifted ID set.}
	\label{T:classic_ood}
	The previous methods of MSP~\cite{baseline}, ODIN~\cite{odin}, EBO score~\cite{energyood}, and MDS~\cite{mahalanobis} reaches a good results on the classic benchmark. However, the value of $p(\bm{x}_n)$ can exceed all the previous methods and achieve a near-perfect result across all the metrics, showing that only taking covariate shift score can completely solve the classic OOD detection benchmark, which, in fact, contradicts the goal of OOD detection. This phenomenon also advocates the significance of the proposed FS-OOD benchmark.}
	\vspace{-5pt}
	\centering
	\resizebox{\textwidth}{!}{
	\begin{tabular}{@{\hskip 8pt}l@{\hskip 6pt}|@{\hskip 6pt}cccccc@{\hskip 6pt}|@{\hskip 6pt}cccccc@{\hskip 6pt}|@{\hskip 6pt}cccccc@{\hskip 8pt}}
		\toprule
		& \multicolumn{6}{c@{\hskip 6pt}|@{\hskip 6pt}}{FPR95~$\downarrow$} 
		& \multicolumn{6}{c@{\hskip 6pt}|@{\hskip 6pt}}{AUROC~$\uparrow$} 
		& \multicolumn{6}{c@{\hskip 6pt}}{AUPR~$\uparrow$} \\ 
		& MSP & ODIN & MDS & EBO & SEM & $p(\bm{x}_n)$ & MSP & ODIN & MDS & EBO & SEM & $p(\bm{x}_n)$ & MSP & ODIN & MDS & EBO & SEM & $p(\bm{x}_n)$ \\
		\midrule
		\multicolumn{16}{l}{\textbf{- DIGITS} (ID: \underline{MNIST})} \vspace{.1cm} \\
		\rowcolor{COLOR_NEAROOD}
		notMNIST 
        & 43.09& 37.70& 44.06& 1.77& 2.64& \textbf{0.78} 
        & 88.77& 89.85& 88.44& 99.67& 99.50& \textbf{99.79} 
        & 75.72& 77.83& 75.97& 99.36& 99.09& \textbf{99.57} \\
        \rowcolor{COLOR_NEAROOD}
	    FashionMNIST 
	   & 2.54& 1.08& 1.05& 0.27& 40.09& \textbf{0.00} 
        & 99.44& 99.70& 99.72& 99.90& 95.02& \textbf{99.94} 
        & 99.64& 99.77& 99.76& 99.94& 97.63& \textbf{99.97} \\
	    \midrule
	    \rowcolor{COLOR_MEAN}
        Mean (Near-OOD) 
        & 20.05& 13.48& 20.54& 2.68& 27.85& \textbf{0.46} 
        & 96.06& 96.97& 95.85& 99.49& 93.85& \textbf{99.78} 
        & 94.07& 94.72& 92.66& 99.40& 93.23& \textbf{99.73} \\
        \midrule
        \rowcolor{COLOR_FAROOD}
        Texture
        & 2.43& 0.94& 0.67& 0.23& 90.69& \textbf{0.02} 
        & 99.34& 99.75& 99.81& 99.93& 77.26& \textbf{99.91} 
        & 99.58& 99.84& 99.84& 99.96& 87.56& \textbf{99.95} \\
        \rowcolor{COLOR_FAROOD}
        CIFAR-10 
        & 7.05& 3.06& 3.18& 0.18& 54.43& \textbf{0.00} 
        & 98.68& 99.31& 99.30& 99.88& 94.19& \textbf{99.97} 
        & 98.72& 99.27& 99.12& 99.88& 95.86& \textbf{99.97} \\
        \rowcolor{COLOR_FAROOD}
        Tiny-ImageNet
        & 6.28& 2.93& 3.13& 0.55& 59.52& \textbf{0.00} 
        & 98.78& 99.36& 99.37& 99.79& 93.70& \textbf{99.96} 
        & 98.78& 99.33& 99.25& 99.79& 95.54& \textbf{99.96} \\
        \rowcolor{COLOR_FAROOD}
        Places365    
        & 9.92& 4.59& 4.12& 0.45& 58.07& \textbf{0.00} 
        & 98.19& 99.06& 99.17& 99.81& 93.82& \textbf{99.96} 
        & 94.87& 97.01& 96.84& 99.42& 91.32& \textbf{99.88} \\
        \midrule
		\rowcolor{COLOR_MEAN}
        Mean (Far-OOD)  
        & 6.45& 2.92& 2.87& 0.36& 53.03& \textbf{0.00} 
        & 98.77& 99.36& 99.39& 99.84& 94.18& \textbf{99.96} 
        & 98.00& 98.84& 98.74& 99.76& 95.09& \textbf{99.94} \\
		\bottomrule
	\end{tabular}}
	\vspace{-5pt}
\end{table*}

\paragraph{Observation Summary}
We summarize the following two take-away messages from the experiments on all three FS-OOD benchmarks:
\textbf{1)} SEM score performs consistently well on near-OOD, which classic output-based methods (\eg, MSP, ODIN, EBO) majorly fail on. The reason can be that output-based methods use too much covariate shift information for OOD detection, which by nature cannot distinguish between covariate-shifted ID and near-OOD.
The proposed SEM score also outperforms the similar feature-based baseline MDS.
\textbf{2)} SEM score sometimes underperforms on far-OOD, with a similar reason that classic OOD detectors use covariate shift to distinguish ID and OOD, which is sometimes sufficient to detect far-OOD samples. Nevertheless, SEM reaches more balanced good results on near-OOD and far-OOD.

\subsection{Results on Classic OOD Detection Setting}
Table~\ref{T:classic_ood} shows the performance on the classic OOD detection benchmark.
The result shows that without the introduction of covariate-shifted ID data, the previous methods reach a near-perfect performance on the classic benchmark, which matches the reported results in their origin papers. 
However, by comparing with Table~\ref{T:main_result}, their performance significantly breakdown when covariate-shifted ID is introduced, showing the fragility of previous methods, and therefore we advocate the more realistic FS-OOD benchmark.
Furthermore, we also report the results that by using the value of $p(\bm{x}_n)$, the score from low-layer feature statistics for detecting covariate shift is shown surprisingly effective on classic OOD benchmark, which exceeds all the previous methods and achieve a near-perfect result across all the metrics. 
This phenomenon shows that only taking covariate shift score can completely solve the classic OOD detection benchmark with MNIST, which, in fact, contradicts the goal of OOD detection. It also advocates the significance of the proposed FS-OOD benchmark.

\subsection{Ablation Study}
\label{S:exp_ablation}
In this section, we validate the effectiveness of the main components that contribute to the proposed SEM score, and also analyze the effects of fine-tuning scheme for source-awareness enhancement. All the experiments in this part are conducted on the DIGITS benchmark.

\begin{table}[t]
\centering
\tabstyle{2.5pt}
	\caption{\textbf{Ablation study on the SEM components.} AUROC is reported for performance evaluation. Several options can be applied to estimate $p(\bm{x}_n)$ and $p(\bm{x})$ in Equation~\ref{eq:final_score}. FS denotes the usage of feature statistics, and FF denotes flattened features. T and L means top-/low-layer feature, \eg, L-FS means low-layer feature statistics.	The results show the effectiveness of our SEM score.}
	\vspace{-5pt}
	\label{T:ablation_sem}
	\resizebox{0.9\linewidth}{!}{
    \begin{tabular}{@{\hskip 4pt}c|ccc|c|cc@{\hskip 4pt}}
    \toprule
    \multirow{2}{*}{\#} & \multicolumn{3}{c|}{$p(\bm{x}_n)$} & \multicolumn{1}{c|}{~$p(\bm{x})$~} & \multirow{2}{*}{~NearOOD} & \multirow{2}{*}{FarOOD~} \\ \cmidrule(lr){2-5}
     & ~T-FS~ & \multicolumn{1}{c}{~L-FF~} & \multicolumn{1}{c|}{~L-FS~} & ~T-FF~ &  &  \\ \midrule
     1 &&&& \cmark         & 87.28 & 60.80\\
     2 & \cmark &&&        & 87.28 & 60.80\\
     3 & \cmark &&& \cmark &   -   &   -  \\
     4 && \cmark &&        & 51.81 & 51.81\\
     5 && \cmark && \cmark & 86.54 & 61.26\\\midrule
     6 &&& \cmark &        & 70.27 & 72.58\\
     7 &&& \cmark & \cmark & \textbf{88.47} & \textbf{69.73}\\
     \bottomrule
    \end{tabular}
	}
	\vspace{-5pt}
\end{table}

\paragraph{Components of SEM}
According to Equation~\ref{eq:decom} in the Section~\ref{S:method}, SEM score can be decomposed by the estimations of $p(\bm{x})$ and $p(\bm{x}_n)$. While our final SEM score uses output flattened features of the CNN model for $p(\bm{x})$ estimation and low-layer feature statistics for $p(\bm{x}_n)$, there are actually several options for the estimation, which is discussed in Table~\ref{T:ablation_sem}.
In this analysis, we set top flattened features as the default usage for $p(\bm{x})$ and only explore $p(\bm{x}_n)$, which is the key part of SEM score.

Exp\#1 shows the result that only uses $p(\bm{x})$ as the final score, which can be interpreted as a simple method using GMM to estimate ID likelihood on the final-layer features. Compared to the MDS result in Table~\ref{T:main_result}, 
this simple method already obtains a better performance on near-OOD.
Notice that we use LeNet-5 on DIGITS, the final-layer features are identical to their feature statistics (ref. Exp\#2). Therefore, everything is cancelled out if $p(\bm{x}_n)$ is top-layer feature statistics (ref. Exp\#3).

\begin{figure}
\centering
\includegraphics[width=0.9\linewidth]{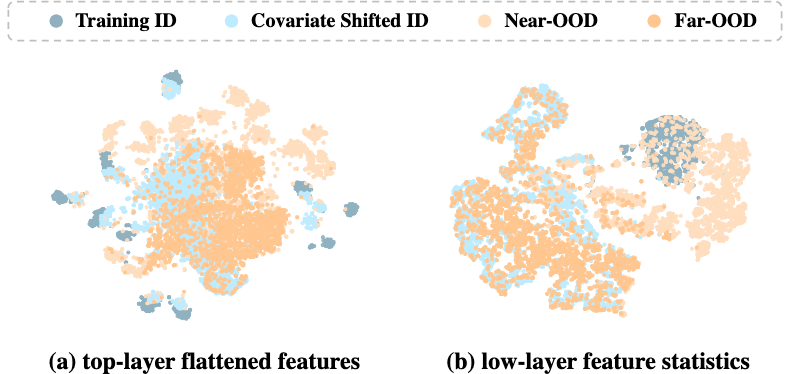}
\vspace{-5pt}
\caption{T-SNE visualization on DIGITS. It suggests that low-layer feature statistics capture non-semantic information, and top-layer features capture both semantic and non-semantic information.}
\vspace{-12pt}
\label{fig:feat_vs_stats}
\end{figure}

Exp\#4 and Exp\#6 shows comparison between using low-layer flattened features (L-FF) and low-layer feature statistics (L-FS) only. 
The performance on detecting covariate-shifted ID shows that both L-FF and L-FS have significant sensitivity to covariate shifts, but with a poor performance on FS-OOD detection. The result indicates that with only the usage of low-level features, the score has a strong correlation to covariate shift but barely contains semantic information, and the feature statistics show the stronger characteristics compared to flattened feature.
This observation indicates our selection of low-level feature statistics for estimating $p(\bm{x}_n)$, which is further supported by the results of Exp\#5 and Exp\#7, and visually illustrated by Figure~\ref{fig:feat_vs_stats}.

\begin{table}[t]
\centering
\tabstyle{2.5pt}
	\caption{\textbf{Ablation study on the fine-tuning scheme for source-awareness enhancement.} AUROC is reported for performance evaluation.
	\#1 reports the performance before fine-tuning.
	$\mathcal{L}_{src}(x)$ means fine-tuning without negative augmented data.
	$\mathcal{L}_{src}(x')$ means only data with negative augmentation is used.
	The results show the effectiveness of each training loss.}
	\label{T:ablation_loss}
	\vspace{-8pt}
	\resizebox{0.9\linewidth}{!}{
	\begin{tabular}{@{\hskip 4pt}l@{\hskip 4pt}|@{\hskip 4pt}ccc@{\hskip 6pt}|@{\hskip 6pt}cc}
	    \toprule
		\# & $\mathcal{L}_{cls}$ & $\mathcal{L}_{src}$($x$) & $\mathcal{L}_{src}$($x'$) & NearOOD & FarOOD \\ 
		\midrule
		1 &&&                   & 83.03 & 56.65 \\
		2 & \cmark &&           & 86.55 & 64.61 \\
		3 & \cmark & \cmark &    & 87.42 & 68.40 \\
		4 & \cmark && \cmark   & 87.27 & 67.92 \\
		\midrule
		5 & \cmark & \cmark & \cmark & \textbf{88.47} & \textbf{69.73} \\
		\bottomrule
	\end{tabular}}
\end{table}

\paragraph{Fine-Tuning Scheme}
Here we evaluate the designed fine-tuning scheme of SEM. 
As elaborated in Section~\ref{sec:method;subsec:enhance}, this learning procedure is designed to enhance the source-aware compactness.
Specifically, a source-awareness enhancement loss $\mathcal{L}_{src}$ is proposed to aggregate the ID training data and separate from the generated negative augmented images at the same time.
Table~\ref{T:ablation_loss} demonstrates the effectiveness of the fine-tuning scheme.
When combining both in-distribution training and negative augmented data training, our framework achieves the best performance.

\begin{table}
    \tabstyle{5pt}
	\caption{\textbf{Hyperparameter Selection of the Number of GMM Components $K$.}
	\label{T:hyperparam_k}
	The result shows that $M=3$ in low-layer statistics and $M=10$ for top-layer features (equal to number of classes) can reach the best results in MNIST benchmark.}
	\vspace{-5pt}
	\centering
	\resizebox{0.9\linewidth}{!}{
    \begin{tabular}{@{\hskip 4pt}c|cc|cc|cc@{\hskip 4pt}}
    \toprule
    \multirow{2}{*}{\#} & \multicolumn{2}{c|}{$p(\bm{x}_n)$} & \multicolumn{2}{c|}{~$p(\bm{x})$~} & \multirow{2}{*}{~NearOOD} & \multirow{2}{*}{FarOOD~} \\ \cmidrule(lr){2-5}
     & ~M=1~ & \multicolumn{1}{c|}{~M=3~} & ~M=10~ & \multicolumn{1}{c|}{~M=20~} &  &  \\ \midrule
     1 &\cmark&&&\cmark        & 86.24 & 64.94\\
     2 &\cmark&&\cmark&        & 85.81 & 60.17\\
     3 &&\cmark&&\cmark        & 84.47 & 63.61\\
     4 &&\cmark&\cmark&        & \textbf{88.47} & \textbf{69.73}\\
     \bottomrule
    \end{tabular}
	}
\end{table}

\paragraph{Hyperparameter of $M$}
Table~\ref{T:hyperparam_k} shows the analysis of hyperparameter $M$. In the DIGITS dataset, $M=3$ leads to a slightly better performance comparing to other choices. Nevertheless, the overall difference among various $M$ is not obvious on near-OOD, showing that the model is robust to the hyperparameter.

\section{Discussion and Conclusion}

Existing OOD detection literature has shown mostly relied on covariate shift even though they are intended to detect semantic shift. This is very effective when test OOD data only come from the far-OOD group---where the covariate shift is large and is further exacerbated by semantic shift, so using covariate shift as a measure to detect OOD fares well. However, when it comes to near-OOD data, especially with covariate-shifted ID (\ie, data experiencing covariate shift but still belonging to the same in-distribution data), current state-of-the-art methods would suffer a significant drop in performance, as shown in the experiments.

We find the gap is caused by a shortcoming in existing evaluation benchmarks: they either exclude covariate-shifted data during testing or treat them as OOD, which is conceptually contradictory with the primary goal that a machine learning model should generalize beyond the training distribution. To fill the gap, we introduce a new problem setting that better matches the design principles of machine learning models: they should be robust in terms of good generalization to covariate-shifted datasets, and trustworthy as they also need to be capable of detecting abnormal semantic shift.

The empirical results suggest that current state-of-the-art methods rely too heavily on covariate shift and hence could easily mis-classify covariate-shifted ID data as OOD data. In contrast, our SEM score function, despite having a simple design, provides a more reliable measure for solving full-spectrum OOD detection.

In fact, to detecting samples with covariate shift, we find that a simple probabilistic model using low-level feature statistics can reach a near-perfect result.

\paragraph{Outlook}
As the OOD detection community getting common awareness of the saturated performance problem of classic OOD benchmarks, several works have taken one-step further towards the more realistic setting and proposed large-scale benchmarks~\cite{wang2022vim,srivastava2022out}. 
However, this paper shows that even under the classic MNIST/CIFAR-scale OOD benchmarks, current OOD methods in fact cannot achieve satisfactory results when the generalization ability is required.
We hope that the future OOD detection works could also consider the generalization capability on covariate-shifted ID data, in parallel to exploring larger-scale models and datasets.

\paragraph{Broader Impacts}
Our research aims to improve the robustness of machine learning systems in terms of the capability to safely handle abnormal data to avoid catastrophic failures. This could have positive impacts on a number of applications, ranging from consumer (e.g., AI-powered mobile phones) to transportation (e.g., autonomous driving) to medical care (e.g., abnormality detection). The new problem setting introduced in the paper includes an important but largely missing element in existing research, namely data experiencing covariate shift but belonging to the same in-distribution classes. We hope the new setting, along with the simple approach based on SEM and the findings presented in the paper, can pave the way for future research for more reliable and practical OOD detection.

{\small
\bibliographystyle{unsrt}
\bibliography{bibtex}}

\begin{thebibliography}{10}

\bibitem{yang2021generalized}
Jingkang Yang, Kaiyang Zhou, Yixuan Li, and Ziwei Liu.
\newblock Generalized out-of-distribution detection: A survey.
\newblock {\em arXiv preprint arXiv:2110.11334}, 2021.

\bibitem{baseline}
Dan Hendrycks and Kevin Gimpel.
\newblock A baseline for detecting misclassified and out-of-distribution
  examples in neural networks.
\newblock In {\em Proceedings of International Conference on Learning
  Representations (ICLR)}, 2017.

\bibitem{odin}
Shiyu Liang, Yixuan Li, and Rayadurgam Srikant.
\newblock Enhancing the reliability of out-of-distribution image detection in
  neural networks.
\newblock In {\em Proceedings of International Conference on Learning
  Representations (ICLR)}, 2017.

\bibitem{energyood}
Weitang Liu, Xiaoyun Wang, John Owens, and Yixuan Li.
\newblock Energy-based out-of-distribution detection.
\newblock In {\em Proceedings of Advances in Neural Information Processing
  Systems (NeurIPS)}, 2020.

\bibitem{mahalanobis}
Kimin Lee, Kibok Lee, Honglak Lee, and Jinwoo Shin.
\newblock A simple unified framework for detecting out-of-distribution samples
  and adversarial attacks.
\newblock In {\em Proceedings of Advances in Neural Information Processing
  Systems (NeurIPS)}, 2018.

\bibitem{likelihood}
Jie Ren, Peter~J Liu, Emily Fertig, Jasper Snoek, Ryan Poplin, Mark Depristo,
  Joshua Dillon, and Balaji Lakshminarayanan.
\newblock Likelihood ratios for out-of-distribution detection.
\newblock In {\em Proceedings of Advances in Neural Information Processing
  Systems (NeurIPS)}, 2019.

\bibitem{duq20icml}
Joost Van~Amersfoort, Lewis Smith, Yee~Whye Teh, and Yarin Gal.
\newblock Uncertainty estimation using a single deep deterministic neural
  network.
\newblock In {\em ICML}, 2020.

\bibitem{gram20icml}
Chandramouli~Shama Sastry and Sageev Oore.
\newblock Detecting out-of-distribution examples with gram matrices.
\newblock In {\em ICML}, 2020.

\bibitem{zhou2021domain}
Kaiyang Zhou, Ziwei Liu, Yu~Qiao, Tao Xiang, and Chen~Change Loy.
\newblock Domain generalization in vision: A survey.
\newblock {\em arXiv preprint arXiv:2103.02503}, 2021.

\bibitem{ming2021impact}
Yifei Ming, Hang Yin, and Yixuan Li.
\newblock On the impact of spurious correlation for out-of-distribution
  detection.
\newblock {\em arXiv preprint arXiv:2109.05642}, 2021.

\bibitem{waic}
Hyunsun Choi, Eric Jang, and Alexander~A Alemi.
\newblock {WAIC}, but why? {Generative} ensembles for robust anomaly detection.
\newblock {\em arXiv preprint arXiv:1810.01392}, 2018.

\bibitem{eloc}
Apoorv Vyas, Nataraj Jammalamadaka, Xia Zhu, Dipankar Das, Bharat Kaul, and
  Theodore~L Willke.
\newblock Out-of-distribution detection using an ensemble of self supervised
  leave-out classifiers.
\newblock In {\em Proceedings of the European Conference on Computer Vision
  (ECCV)}, 2018.

\bibitem{S}
Joan Serr{\`a}, David {\'A}lvarez, Vicen{\c{c}} G{\'o}mez, Olga Slizovskaia,
  Jos{\'e}~F N{\'u}{\~n}ez, and Jordi Luque.
\newblock Input complexity and out-of-distribution detection with
  likelihood-based generative models.
\newblock In {\em Proceedings of International Conference on Learning
  Representations (ICLR)}, 2020.

\bibitem{hendrycks18oe}
Dan Hendrycks, Mantas Mazeika, and Thomas Dietterich.
\newblock Deep anomaly detection with outlier exposure.
\newblock In {\em Proceedings of International Conference on Learning
  Representations (ICLR)}, 2019.

\bibitem{backgroundsample}
Yi~Li and Nuno Vasconcelos.
\newblock Background data resampling for outlier-aware classification.
\newblock In {\em Proceedings of the IEEE Conference on Computer Vision and
  Pattern Recognition (CVPR)}, 2020.

\bibitem{yang2021scood}
Jingkang Yang, Haoqi Wang, Litong Feng, Xiaopeng Yan, Huabin Zheng, Wayne
  Zhang, and Ziwei Liu.
\newblock Semantically coherent out-of-distribution detection.
\newblock In {\em Proceedings of the IEEE International Conference on Computer
  Vision (ICCV)}, 2021.

\bibitem{mixup}
Hongyi Zhang, Moustapha Cisse, Yann~N Dauphin, and David Lopez-Paz.
\newblock mixup: Beyond empirical risk minimization.
\newblock In {\em Proceedings of International Conference on Learning
  Representations (ICLR)}, 2018.

\bibitem{lin2021domain}
Chuang Lin, Zehuan Yuan, Sicheng Zhao, Peize Sun, Changhu Wang, and Jianfei
  Cai.
\newblock Domain-invariant disentangled network for generalizable object
  detection.
\newblock In {\em Proceedings of the IEEE International Conference on Computer
  Vision (ICCV)}, 2021.

\bibitem{peng2019domain}
Xingchao Peng, Zijun Huang, Ximeng Sun, and Kate Saenko.
\newblock Domain agnostic learning with disentangled representations.
\newblock In {\em Proceedings of International Conference on Machine Learning
  (ICML)}, 2019.

\bibitem{mixstyle}
Kaiyang Zhou, Yongxin Yang, Yu~Qiao, and Tao Xiang.
\newblock Domain generalization with mixstyle.
\newblock In {\em Proceedings of International Conference on Learning
  Representations (ICLR)}, 2021.

\bibitem{adain}
Xun Huang and Serge Belongie.
\newblock Arbitrary style transfer in real-time with adaptive instance
  normalization.
\newblock In {\em Proceedings of the IEEE Conference on Computer Vision and
  Pattern Recognition (CVPR)}, 2017.

\bibitem{bengio2013representation}
Yoshua Bengio, Aaron Courville, and Pascal Vincent.
\newblock Representation learning: A review and new perspectives.
\newblock {\em IEEE Transactions on Pattern Analysis and Machine Intelligence
  (TPAMI)}, 2013.

\bibitem{sinha2021negative}
Abhishek Sinha, Kumar Ayush, Jiaming Song, Burak Uzkent, Hongxia Jin, and
  Stefano Ermon.
\newblock Negative data augmentation.
\newblock In {\em Proceedings of International Conference on Learning
  Representations (ICLR)}, 2021.

\bibitem{mnist}
Yann LeCun, Corinna Cortes, and Christopher~JC Burges.
\newblock The mnist database of handwritten digits, 1998.
\newblock \url{http://yann. lecun. com/exdb/mnist}, 1998.

\bibitem{svhn}
Yuval Netzer, Tao Wang, Adam Coates, Alessandro Bissacco, Bo~Wu, and Andrew~Y
  Ng.
\newblock Reading digits in natural images with unsupervised feature learning.
\newblock In {\em Proceedings of NIPS Workshop on Deep Learning and
  Unsupervised Feature Learning}, 2011.

\bibitem{usps}
Jonathan~J. Hull.
\newblock A database for handwritten text recognition research.
\newblock {\em IEEE Transactions on pattern analysis and machine intelligence},
  16(5):550--554, 1994.

\bibitem{notmnist}
Yaroslav Bulatov.
\newblock {NotMNIST} dataset.
\newblock \url{http://yaroslavvb.blogspot.com/2011/09/notmnist-dataset.html},
  2011.

\bibitem{fashionmnist}
Han Xiao, Kashif Rasul, and Roland Vollgraf.
\newblock Fashion-mnist: a novel image dataset for benchmarking machine
  learning algorithms.
\newblock {\em arXiv preprint arXiv:1708.07747}, 2017.

\bibitem{texture}
Mircea Cimpoi, Subhransu Maji, Iasonas Kokkinos, Sammy Mohamed, and Andrea
  Vedaldi.
\newblock Describing textures in the wild.
\newblock In {\em Proceedings of the IEEE Conference on Computer Vision and
  Pattern Recognition (CVPR)}, 2014.

\bibitem{cifar}
Alex Krizhevsky, Geoffrey Hinton, et~al.
\newblock Learning multiple layers of features from tiny images.
\newblock {\em Citeseer}, 2009.

\bibitem{imagenet}
Olga Russakovsky, Jia Deng, Hao Su, Jonathan Krause, Sanjeev Satheesh, Sean Ma,
  Zhiheng Huang, Andrej Karpathy, Aditya Khosla, Michael Bernstein, et~al.
\newblock Imagenet large-scale visual recognition challenge.
\newblock {\em International Journal of Computer Vision (IJCV)}, 2015.

\bibitem{places365}
Bolei Zhou, Agata Lapedriza, Aditya Khosla, Aude Oliva, and Antonio Torralba.
\newblock Places: A 10 million image database for scene recognition.
\newblock {\em IEEE Transactions on Pattern Analysis and Machine Intelligence
  (TPAMI)}, 2017.

\bibitem{santa2021public}
Beatriz~Garcia Santa~Cruz, Mat{\'\i}as~Nicol{\'a}s Bossa, Jan S{\"o}lter, and
  Andreas~Dominik Husch.
\newblock Public covid-19 x-ray datasets and their impact on model bias--a
  systematic review of a significant problem.
\newblock {\em Medical image analysis}, 2021.

\bibitem{vaya2020bimcv}
Maria de la~Iglesia Vay{\'a}, Jose~Manuel Saborit, Joaquim~Angel Montell,
  Antonio Pertusa, Aurelia Bustos, Miguel Cazorla, Joaquin Galant, Xavier
  Barber, Domingo Orozco-Beltr{\'a}n, Francisco Garc{\'\i}a-Garc{\'\i}a, et~al.
\newblock Bimcv covid-19+: a large annotated dataset of rx and ct images from
  covid-19 patients.
\newblock {\em arXiv preprint arXiv:2006.01174}, 2020.

\bibitem{Wang2020}
Linda Wang, Zhong~Qiu Lin, and Alexander Wong.
\newblock Covid-net: a tailored deep convolutional neural network design for
  detection of covid-19 cases from chest x-ray images.
\newblock {\em Scientific Reports}, 2020.

\bibitem{winther12275009covid}
Hinrich~B Winther, Hans Laser, Svetlana Gerbel, Sabine~K Maschke, Jan~B
  Hinrichs, Jens Vogel-Claussen, Frank~K Wacker, Marius~M H{\"o}per, and
  Bernhard~C Meyer.
\newblock Covid-19 image repository. 2020.
\newblock {\em URL https://figshare.
  com/articles/COVID-19\_Image\_Repository/12275009}.

\bibitem{boneage}
RSNA.
\newblock {RSNA Pediatric Bone Age Challenge} (2017), 2017.

\bibitem{covidct}
Xingyi Yang, Xuehai He, Jinyu Zhao, Yichen Zhang, Shanghang Zhang, and Pengtao
  Xie.
\newblock Covid-ct-dataset: a ct scan dataset about covid-19.
\newblock {\em arXiv preprint arXiv:2003.13865}, 2020.

\bibitem{lenet}
Yann LeCun et~al.
\newblock Lenet-5, convolutional neural networks.
\newblock \url{http://yann. lecun. com/exdb/lenet}, 2015.

\bibitem{resnet}
Kaiming He, Xiangyu Zhang, Shaoqing Ren, and Jian Sun.
\newblock Deep residual learning for image recognition.
\newblock In {\em Proceedings of the IEEE Conference on Computer Vision and
  Pattern Recognition (CVPR)}, 2016.

\bibitem{wang2022vim}
Haoqi Wang, Zhizhong Li, Litong Feng, and Wayne Zhang.
\newblock Vim: Out-of-distribution with virtual-logit matching.
\newblock In {\em Proceedings of the IEEE Conference on Computer Vision and
  Pattern Recognition (CVPR)}, 2022.

\bibitem{srivastava2022out}
Anugya Srivastava, Shriya Jain, and Mugdha Thigle.
\newblock Out of distribution detection on imagenet-o.
\newblock {\em arXiv preprint arXiv:2201.09352}, 2022.

\end{thebibliography}

\end{document}